# BLC: Private Matrix Factorization Recommenders via Automatic Group Learning

Alessandro Checco, School of Computer Science and Statistics, Trinity College, Dublin, Ireland.
Giuseppe Bianchi, School of Engineering, Università di Roma "Tor Vergata", Rome, Italy.
Douglas J. Leith, School of Computer Science and Statistics, Trinity College, Dublin, Ireland.

We propose a privacy-enhanced matrix factorization recommender that exploits the fact that users can often be grouped together by interest. This allows a form of "hiding in the crowd" privacy. We introduce a novel matrix factorization approach suited to making recommendations in a shared group (or nym) setting and the BLC algorithm for carrying out this matrix factorization in a privacy-enhanced manner. We demonstrate that the increased privacy does not come at the cost of reduced recommendation accuracy.

## 1. INTRODUCTION

In a classical recommender system a set of users rate a set of items and these ratings are then used to predict user ratings for those items which they have not yet rated, see Figure 1. The ratings supplied by each user are tagged with the user identity, e.g. via a browser cookie, and so the ratings are individually identifiable. This raises obvious privacy concerns, whereby an attacker observing the item ratings submitted by each user may learn information which the user does not wish to disclose. Such an attacker might observe the rating by sniffing the network path between a user and the recommender system or, more likely, may be the recommender system itself so that encryption of the network traffic does not constitute a defence.

It is well known that users can often be grouped together by e.g. interest in sport, type of movie or choice of partner, and indeed this underpins most advertising campaigns. While such grouping is often carried out manually, there has also been interest in automated inference of abstract groups and it is such unsupervised automated inference of abstract groups that we pursue in the present paper. Namely, we start with the observation that the existence of abstract group structure raises the potential to use a group identity when submitting ratings to a recommender system instead of an individual user identity, and in this way provide a form of "hiding in the crowd" privacy. One key challenge, which we fully resolve here, is to provide accurate recommendations to a user without requiring the user to disclose their personal ratings to the recommender system. This is essential to protect against the type of privacy disclosure attack considered here. Another challenge is the selection of appropriate abstract groups that yield good prediction accuracy while being shared by a sufficiently large number of users so as to provide a degree of privacy. In this paper we demonstrate that this is indeed possible for interesting, real data sets and consequently that the gain in privacy need not come at the cost of reduced recommender prediction accuracy.

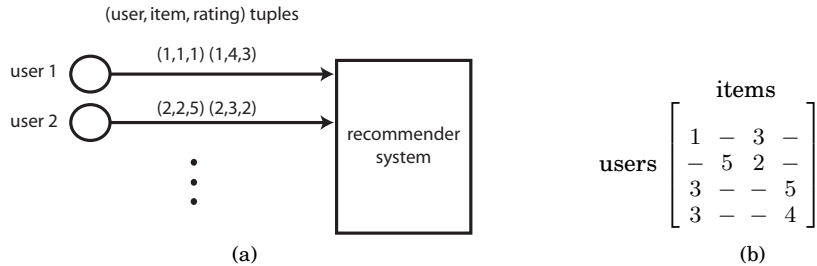

Fig. 1. Illustrating classical recommender system setup. Each user submits ratings for a set of items, where this set may be different for each user, the ratings being tagged with the user identity, see left-hand plot (a). The submitted (user,item rating) pairs can be gathered into a matrix $R$, see right-hand plot (b), and the aim of the system is to predict the missing values, indicated by $-$.

In more detail, we introduce a set of artificial pseudo-identities, referred to as *proxynyms* (or, in short, nyms), with which users access the recommending service. These nyms differ from ordinary pseudonyms in that the same nym is shared by many users. From the point of view of the system, therefore, it receives a sequence of item ratings from each of $p$ nyms rather than from $n$ users. To make this setup work two main tasks must be addressed: (i) each user must privately select an appropriate nym when submitting ratings and (ii) the received collection of nym-item ratings must be capable of being used to make accurate predictions of the user ratings. Observe that these two tasks are coupled and must be solved in a joint manner i.e. the set of users selecting a nym must share enough in common to permit accurate predictions to be made when only nym-item ratings are observed. Further, unlike in the classical setup the same item may often be rated multiple times by a nym, since multiple users share the nym, and the nyms are not all equally "important" since one may be shared by many more users than another.

In order to make recommendations we build on existing matrix factorization approaches in which the matrix $R$ of user-item ratings (see Figure 1(b)) is decomposed as a product $U^T V$, where the row dimension of matrix $U$ and matrix $V$ is much less than the numbers of users and items. The user-nym selection and nym-item matrix factorisation tasks can be viewed as decomposing the $n \times m$ user-item rating matrix $R$ as $P^T \tilde{U}^T V$ where $n \times p$ matrix $P^T$ maps from users to nyms. The elements of $P$ are $\{0, 1\}$ valued and $P$ is column stochastic (the columns sum to one) so that each user is a member of a single nym (extension to multiple nym membership is of course possible). One of our main contributions is the observation that by use of the BLC algorithm it is in fact possible to carry out this $P$, $\tilde{U}$, $V$ decomposition in a privacy-enhancing manner, without the need for sophisticated cryptographic methods.

At the cost of introducing a more complex factorisation task, we find that prediction performance competitive with the state of the art can be obtained using this approach, especially when the user rating matrix $R$ is sparse (as it usually is). That is, the increased privacy does not come at the cost of reduced recommendation accuracy. Indeed, we show that it can yield improved accuracy over classical matrix factorisation. As already noted, this is perhaps unsurprising since, when nyms are chosen appropriately, users sharing a nym can usefully leverage their shared ratings/preferences in a more direct way than in classical collaborative filtering, where such leverage can become apparent only when the dimension of the latent space is very small. Further, we find that the projection $P$ provides a useful form of regularisation that greatly reduces the sensitivity of performance to the choice of hyperparameters. While more computationally demanding that the standard matrix factorisation approach, BLC is still efficient and highly scalable.

In summary, the main contributions of the present paper are as follows: (i) a novel matrix factorization approach suited to making recommendations in a shared nym setting, (ii) the BLC algorithm for carrying out this matrix factorization in a privacy-enhanced manner and (iii) its performance evaluation using a mix of both real and synthetic data sets. Although not the focus of the present paper, we also note that a relevant feature of our approach is its potential for backward compatibility with many existing recommender systems, and so its potential suitability for incremental roll-out.

## 2. RELATED WORK

The potential for privacy concerns in recommender systems is well known, e.g. see [Lam and Riedl 2004; Calandrino et al. 2011; Shyong et al. 2006; Ding et al. 2010; Narayanan and Shmatikov 2006] and references therein.

Many existing approaches to making private recommendations are based on encryption and multi-party computation [Guha et al. 2011; Li et al. 2011; Aïmeur et al. 2008; Nikolaenko et al. 2013; Canny 2002] and are essentially clean-slate designs (not backward compatible with existing recommender systems). In [Nikolaenko et al. 2013], a trusted cryptographic service provider performs an encrypted version of the matrix factorization task used to calculate recommendations. In [Aïmeur et al. 2008], the user data is split between the merchant and a semi-trusted third party; if these two actors do not collude, the cryptographic system proposed allows user recommendations to be calculated without exposing private information. In [Li et al. 2011], content ratings and item similarity data are determined via distributed cryptographic multi-party computation, recommendations are then generated based on interest groups and further personalised at each user's local machine. In the special case where the recommender system's task is to select which adverts to display, another approach [Guha et al. 2011] is for the web system to supply a large set of possible adverts to the user and for the user's browser to then privately make the recommendation decision as to which adverts are actually displayed. In [Canny 2002], a privacy-preserving scheme for a Singular Value Decomposition (SVD) based CF algorithm is introduced.

Another strand of work on privacy-enhanced recommender systems improves privacy by perturbing ratings with noise. Initially proposed by [Agrawal and Srikant 2000] it is applied to collaborative filtering by e.g. [Huang et al. 2005; Kargupta et al. 2003; Polat and Du 2005].

Perhaps the closest to the present work is that of [Shokra et al. 2009] who propose a peer to peer system for submitting aggregated ratings to a central recommender system. In [Nandi et al. 2011] a middleware framework for *supervised* aggregation is discussed but recommender performance is not considered. In [Narayanan and Shmatikov 2006], a technique the distributed aggregation of online profiles is presented, but it requires a certain level of trust and co-ordination between users, and being peer-to-peer, privacy and recommendation quality strongly depend on the user connectivity.

Clustering of users in recommender systems in general is not new. In [Ungar and Foster 1998; Xue et al. 2005; Hofmann 2004] a range of techniques (K-means and Gibbs sampling, a cluster-based smoothing system, and a cluster-based latent semantic model respectively) are proposed which cluster users and items in an unsupervised manner. However, these are not suited to implementation in a distributed and private manner. Clustering with matrix factorization methods has received much less attention to date, with the notable exception of [Ding et al. 2006], and subsequent work building on this, in which an orthogonal non-negative matrix tri-factorization method is introduced. In [Zhu et al. 2014] it is observed that users with a few ratings can leverage the presence of users with many ratings to improve prediction accuracy, and in [Xin and Jaakkola 2014] considers a semi-private variant where private users can leverage the presence of public users with many ratings.

Finally, even if BLC deals with groups, it differs (in scope and technical construction) from traditional group recommender systems. Such systems were originally conceived [O'Connor et al. 2001] for social contexts in which people operate in groups (e.g. deciding which movie to see together, or which restaurant to attend). Their goal is to provide suggestions suitable to *all* members of the group. As comprehensively surveyed in [Boratto and Carta 2011; Kompan and Bielikova 2013], existing work differs in the use of user preferences and/or social information, and in how such information is accounted when making recommendations for groups (or even settling disagreements between group members [Shang et al. 2011]). BLC is *not* a group recommender in the meaning above; it is an *individual* matrix factorization recommender that additionally *clusters* users into groups to improve the recommender's accuracy. In this sense, and putting aside our main privacy goal, BLC may be considered closer, in its scope, to content-boosted [Nguyen and Zhu 2013] or category/topic-based [Zhou et al. 2014; Wang et al. 2012] recommenders, and to a greater extent closer to information matching approaches [Gorla et al. 2013], which show that (probabilistic) matrix factorization approaches can be improved by adding information on content categories or user groups (which are built using user attributes or features). However, we stress that, unlike these works, we *do not exploit any exogenous information* such as user preferences of item categories, and we do not boost the recommender with an a priori computed group structure, but we try to *learn* groups through the own users' ratings, and we do this *while* performing the matrix factorization.

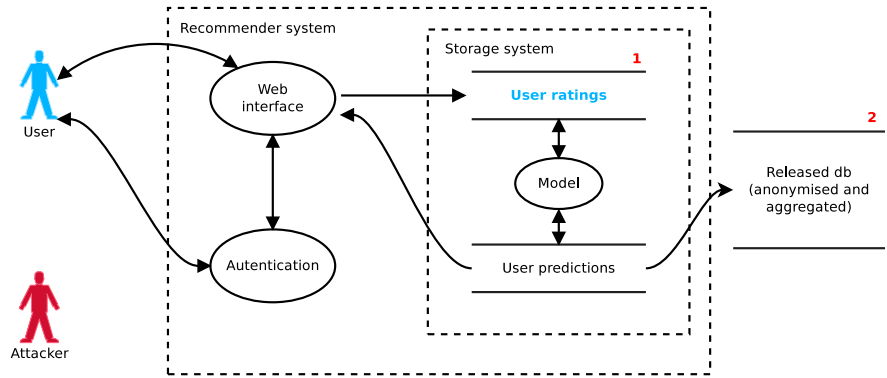

Fig. 2. Data flow diagram of the system scenario with two possible points of attack.

## 3. THREAT MODEL

The privacy disclosure scenario we consider arises naturally from attacks which have already occurred on production recommender systems [Narayanan and Shmatikov 2006]. Figure 2 illustrates the setup. A user sends encrypted ratings and receives recommendations through an authentication based system. The user ratings are kept in the system storage, where more manipulation is needed to generate the recommendations. The recommender system can also decide to release an anonymised/aggregated version of the database. We imagine two main types of attack/threats[1]:

(1) The attacker has access to a leaked version of the user ratings database held by the recommender system.
(2) The recommender system can decide to release an anonymised version of the user ratings database.

It is important to notice that the sensitive asset here is the stream of pairs of unique IDs and ratings. Knowledge of these potentially places two sorts of information at risk: (i) the mapping from ID to true user identity (which may be of interest in its own right) and (ii) the ratings submitted by the user being targetted for attack. When a

---

[1]Note that we do not consider attacks that aim to compromise the user client/browser or attacks against the encryption on the link between user and server, both of which are already much studied.

A:5

single user is associated with each ID, then since the rating profile associated with each ID is often unique there is a risk that the association of a user with an ID may be inferred by de-anonymisation using external data [Datta et al. 2012; Aggarwal 2005]. In this case both the mapping from ID to user identity and the ratings submitted by a user are then simultaneously compromised.

With this kind of attack in mind, our interest is instead in the situation where multiple users are associated with each ID. This changes matters significantly. Firstly, consider an attack that aims to learn the mapping from ID to true user identity. When an ID is shared by $k$ users then the collection of (item, rating) pairs associated with that ID is now a mixture of ratings from the $k$ different users. Provided the $k$ users sharing an ID submit sufficiently different sets of (item, rating) pairs then this attack can be expected to involve a harder inference task than when an ID is used by a single user. This is because the (item, rating) pairs submitted by the other users sharing the ID now act as "noise" that tends to mask the pattern of (item, rating) pairs submitted by the user being targetted. Of course, if the users sharing an ID submit sets of rating which are too similar to each other then the protection against de-anonymisation provided by this direct mixing mechanism may be reduced, in which case additional measures can be envisioned e.g. inserting dummy ratings or other "noise" to increase diversity.

Secondly, when multiple users are associated with each ID then linking a user to an ID no longer immediately reveals that user's ratings, or even the more limited information consisting of the set of items rated by that user[2]. To see this, suppose for a particular item that only $n$ of the $k$ users sharing an ID have rated that item. Then an attacker does not learn whether the user of interest belongs to the set of $n$ users who have rated that item or the set of $k - n$ users who have not. It is only when all users sharing an ID rate a particular item, i.e. $n$ equals $k$ for that item, that the attacker learns that the target user has rated that item. Even in that case the attacker does not learn the rating submitted by the user unless all users sharing the ID also rate the item identically. This direct "hiding in the crowd" mechanism might be further strengthened, if needed, by users additionally submitting a number of automated/dummy (item,rating) pairs selected at random – then, even when an item rated by all users sharing an ID, any individual user has a degree of plausible deniability in that they can claim that their rating for that item was an automated/dummy one.

It can be seen that associating multiple users with each ID in a recommender system not only potentially makes the considered de-anonymisation attacks harder to carry out but also creates the foundation upon which a number of still stronger mechanisms for enhancing privacy can be built. Supporting shared IDs, which we refer to as nyms, while still providing useful recommendations is also the main technical challenge (inserting dummy ratings etc is relatively straightforward) and in the following sections we demonstrate that this can indeed be achieved.

### 3.1. Dishonest Users and Sybil Attacks

Attacks by dishonest users who submit false ratings in an attempt to manipulate the recommendations made by the system are outwith the scope of the present paper. Of course this is an important challenge for all recommender systems, but it is not specific to the approach presented here. That said, the use of shared IDs/nyms does potentially facilitate Sybil attacks and so we briefly describe one mechanism, based on the work of [Chaum et al. 1990], by which such attacks can be disrupted while making use of nyms. In summary, each user mints a number of session tokens (with associated serial number), blinds them with a secret blinding factor and forwards them to the recommender system through a non-secure channel. The number of tokens available to a user is limited e.g. by requiring users to authenticate or make payment to the service in order to forward a token, or perhaps by limiting the number of tokens allowed within a certain time window. Note that during this phase the user might be identified to the system, e.g. to make a payment. The system then signs the tokens with its private key, without knowledge of the serial number associated with the tokens. On receiving the signed tokens back from the recommender system, the user can remove the blinding factor and use the tokens to submit ratings to the system anonymously. Double use of tokens is prevented by the system maintaining a database of the serial numbers of all tokens that have been issued.

---

[2]While an attacker might potentially also be interested in which items have *not* been rated by a user, in a recommender system the ratings by a user are usually extremely sparse, i.e. the number of available items is much larger than the number of ratings submitted by any individual user. Absence of a rating is therefore typically much less informative than the presence of a rating.

## 4. MAPPING USERS TO NYMS

BLC decomposes the user-rating matrix $R$ as $P^T \tilde{U}^T V$. The primary challenges are (i) finding an appropriate assignment of users to nyms i.e. the selection of matrix $P$ and (ii) the calculation of this assignment in privacy-enhanced manner.

### 4.1. Statistical Model

As in classic matrix factorization approaches, let vector $U_u \in \mathbb{R}^d$ associated with user $u$ capture the users preferences in the latent feature space of dimension $d$, and gather these vectors together to form matrix $U \in \mathbb{R}^{d \times n}$. Similarly, let vector $V_v \in \mathbb{R}^d$ associated with resource $v$ capture its features in the latent space and we gather these vectors together to form matrix $V \in \mathbb{R}^{d \times m}$. The rating of item $v$ by user $u$ is Gaussian random variable with mean $U_u^T V_v$ and the matrix of ratings of all items by all users is therefore a Gaussian random variable with mean $U^T V$.

Let us now depart from the usual setup by further assuming that users belong to distinct groups.

ASSUMPTION 1 (NYM DECOMPOSITION). *Matrix $U$ can be decomposed as $U = \tilde{U} P$ where $\tilde{U} \in \mathbb{R}^{d \times p}$ and $P \in \mathbb{R}^{p \times n}$.*

We can think of vector $\tilde{U}_g \in \mathbb{R}^d$, $g = 1, \cdots, p$ as the preference vector associated with a group of users. The preference vectors of the users belonging to that group can then be viewed as random variables with mean $\tilde{U}_g$. We refer to each group $g$ as a *nym*. Matrix $P$ then maps from nyms to users, with $u$'th column $P_u \in \mathbb{R}^p$ of $P$ defining the mapping from the nyms to user $u$.

The rating supplied by user $u$ for resource $v$ is a Gaussian random variable $X_{R_{uv}} \in \mathbb{R}$ with mean $U_u^T V_v = (\tilde{U} P_u)^T V_v$ and variance $\sigma^2$. That is,

$$Prob(X_{R_{uv}} = R_{uv} | \tilde{U}, V; P) \sim e^{-\phi_{uv}(R_{uv})/\sigma^2}$$

where $\phi_{uv}(R_{uv}) := (R_{uv} - P_u^T \tilde{U}^T V_v)^2$. Gathering the user ratings into random matrix $X_Z \in \mathbb{R}^{n \times m}$, let $\mathcal{O} \subset \{1, \cdots, n\} \times \{1, \cdots, m\}$ denote the set of user-resource rating pairs that are contributed by the users. Letting $\mathcal{Z}_\mathcal{O} = \{X_{R_{uv}}, (u,v) \in \mathcal{O}\}$ and $\mathcal{R}_\mathcal{O} = \{R_{uv}, (u,v) \in \mathcal{O}\}$, the conditional distribution over these observed ratings is,

$$Prob(\mathcal{Z}_\mathcal{O} = \mathcal{R}_\mathcal{O} | \tilde{U}, V; P) := p(\mathcal{R}_\mathcal{O} | \tilde{U}, V; P) \sim \prod_{(u,v) \in \mathcal{O}} e^{-\phi_{uv}(R_{uv})/\sigma^2}$$

The posterior distribution is

$$p(\tilde{U}, V | \mathcal{R}_\mathcal{O}; P) = \frac{p(\mathcal{R}_\mathcal{O}|\tilde{U}, V; P)p(\tilde{U})p(V)}{p(\mathcal{R}_\mathcal{O})}$$

and so the log-posterior is

$$\log p(\tilde{U}, V | \mathcal{R}_\mathcal{O}; P) = -\frac{1}{\sigma^2} \sum_{(u,v)\in\mathcal{O}} \phi_{uv}(R_{uv}) - \frac{1}{\sigma_{\tilde{U}}^2} tr \tilde{U}^T \tilde{U} - \frac{1}{\sigma_V^2} tr V^T V + C \qquad (1)$$

where $C$ is a normalising constant and we have assumed Gaussian priors for $\tilde{U}$ and $V$ with zero mean and variance $\sigma_{\tilde{U}}^2$ and $\sigma_{\tilde{V}}^2$, respectively.

### 4.2. Private Nym-Based Matrix Factorization

Use of nyms aside, the log-posterior expression (1) is of the standard form widely used in the matrix factorization recommender literature. To find the matrices $\tilde{U}$, $V$ maximising $\log p(\tilde{U}, V | \mathcal{R}_\mathcal{O}; P)$ requires knowledge of the rating $R_{uv}$ made by each user and when this can be linked to the user identity, e.g. via a cookie, then this is evidently non-private. However, when the user ratings possess a nym structure we have the following key observation:

LEMMA 4.1 (EQUIVALENCE OF LOG-POSTERIOR). *Let*
$\mathcal{U}(v) = \{u : (u, v) \in \mathcal{O}\}$ *be the set of users rating item $v$, and $\mathcal{V} = \{v : (u,v) \in \mathcal{O}, u \in \{1, \cdots, n\}\}$ the set of items for which ratings are observed. Suppose matrix* $\Lambda(v) := \sum_{u \in \mathcal{U}(v)} P_u P_u^T$ *is non-singular. Then matrices $\tilde{U}$, $V$ maximise log-posterior (1) if and only if they maximise*
$\log \tilde{p}(\tilde{U}, V | \mathcal{R}_\mathcal{O}; P)$, *i.e.*

$$\arg\max_{\tilde{U},V} \log p(\tilde{U}, V|\mathcal{R}_\mathcal{O}; P) = \arg\max_{\tilde{U},V} \log \tilde{p}(\tilde{U}, V|\mathcal{R}_\mathcal{O}; P) \qquad (2)$$

*where*

$$\log \tilde{p}(\tilde{U}, V|\mathcal{R}_\mathcal{O}; P) = -\frac{1}{\sigma^2} \sum_{v \in \mathcal{V}} (\tilde{R}_v - \tilde{U}^T V_v)^T \Lambda(v)(\tilde{R}_v - \tilde{U}^T V_v) - \frac{1}{\sigma_{\tilde{U}}^2} tr\tilde{U}^T \tilde{U} - \frac{1}{\sigma_V^2} tr V^T V \qquad (3)$$

*and* $\tilde{R}_v = \Lambda^{-1}(v) \sum_{u \in \mathcal{U}(v)} R_{uv} P_u$.

PROOF. Letting $\psi := -\frac{1}{\sigma_{\tilde{U}}^2} tr \tilde{U}^T \tilde{U} - \frac{1}{\sigma_V^2} tr V^T V$, then

$$\arg\max_{\tilde{U},V} \log p(\tilde{U}, V|\mathcal{R}_\mathcal{O}; P) = \arg\max_{\tilde{U},V} \sum_{(u,v)\in\mathcal{O}} (R_{uv} - (\tilde{U}P_u)^T V_v)^2 + \psi \qquad (4)$$

$$\stackrel{(a)}{=} \arg\max_{\tilde{U},V} \sum_{(u,v)\in\mathcal{O}} -2R_{uv}(\tilde{U}P_u)^T V_v + ((\tilde{U}P_u)^T V_v)^2 + \psi$$

$$= \arg\max_{\tilde{U},V} \sum_{v\in\mathcal{V}} \sum_{u\in\mathcal{U}(v)} -2R_{uv} P_u^T \tilde{U}^T V_v + V_v^T \tilde{U} P_u P_u^T \tilde{U}^T V_v + \psi,$$

where $(a)$ follows from the fact that $R_{uv}^2$ does not depend on $\tilde{U}$ or $V$. Hence, letting $\hat{R}_v := \sum_{u \in \mathcal{U}(v)} R_{uv} P_u$ and using the fact that $\Lambda(v)$ is non-singular, we have that

$$\arg\max_{\tilde{U},V} \log p(\tilde{U}, V|\mathcal{R}_\mathcal{O}; P) = \arg\max_{\tilde{U},V} \sum_{v\in\mathcal{V}} -2\hat{R}_v^T \tilde{U}^T V_v + V_v^T \tilde{U} \sum_{u\in\mathcal{U}(v)} P_u P_u^T \tilde{U}^T V_v + \psi$$

$$= \arg\max_{\tilde{U},V} \sum_{v\in\mathcal{V}} -2\hat{R}_v^T \tilde{U}^T V_v + (\tilde{U}^T V_v)^T \Lambda(v) \tilde{U}^T V_v + \psi \qquad (5)$$

$$\stackrel{(b)}{=} \arg\max_{\tilde{U},V} \sum_{v \in \mathcal{V}} \hat{R}_v^T (\Lambda(v)^{-1})^T \hat{R}_v - 2\hat{R}_v^T \tilde{U}^T V_v$$
$$+ (\tilde{U}^T V_v)^T \Lambda(v) \tilde{U}^T V_v + \psi \quad (6)$$
$$= \arg\max_{\tilde{U},V} \log \tilde{p}(\tilde{U}, V | \mathcal{R}_\mathcal{O}; P), \quad (7)$$

where $(b)$ follows from the fact that $\hat{R}_v^T (\Lambda(v)^{-1})^T \hat{R}_v$ does not depend on $\tilde{U}$ or $V$. □

The condition that $\Lambda(v)$ is non-singular requires that the set of users rating item $v$ is non-degenerate in the sense that every nym is used by at least one user (no zero rows in $\Lambda(v)$) and the vectors of user assignments to each nym are linearly independent (the nyms are distinct).

By Lemma 4.1, matrices $\tilde{U}$, $V$ maximising $\log \tilde{p}(\tilde{U}, V | \mathcal{R}_\mathcal{O}; P)$ in (3) also maximises the log-posterior (1). As we will discuss in more detail shortly, the importance of Lemma 4.1 is that finding matrices $\tilde{U}$, $V$ maximising (3) only requires observations of nym ratings $\tilde{R}_v$ and matrices $\Lambda(v)$, $\forall v$. That is, does *not* require that individual user ratings are observed by the recommender system.

Of particular interest is the special case where users are each assigned to a single nym, and so the columns $P_u$ of matrix $P$ have a single non-zero element (corresponding to the nym chosen) equal to 1. In this case matrix $\Lambda(v)$ is a diagonal matrix with element $\Lambda(v)_{gg}$ equal to the number of users who rate item $v$ using nym $g$ and nym rating vector $\tilde{R}_v$ has element $g$ equal to the average of the ratings for element $v$ by users in nym $g$. Using Lemma 4.1 we then have:

THEOREM 4.2 (PRIVATE NYM FACTORIZATION). *Suppose $\Lambda(v)$ is diagonal. Let $\bar{U}$ and $V$ satisfy*

$$\tilde{U}_g^T = \sum_{v \in \mathcal{V}} \Lambda(v)_{gg} \tilde{R}_{gv} V_v^T \left( \frac{\sigma^2}{\sigma_{\tilde{U}}^2} I + \sum_{w \in \mathcal{V}} \Lambda(w)_{gg} V_w V_w^T \right)^{-1} \quad (8)$$

$$V_v^T = \tilde{R}_v^T \Lambda(v) \tilde{U}^T \left( \frac{\sigma^2}{\sigma_V^2} I + \tilde{U} \Lambda(v) \tilde{U}^T \right)^{-1}. \quad (9)$$

*Then $\tilde{U}, V$ is a stationary point of* (3).

PROOF. Recall that $tr V^T V = \sum_v \sum_i (V_{vi})^2$ and so the derivative with respect to vector $V_v$ is $2V_v^T$. Observe also that the derivative of $(\tilde{R}_v - \tilde{U}^T V_v)^T \Lambda(v)(\tilde{R}_v - \tilde{U}^T V_v)$ with respect to vector $V_v$ is $-2\tilde{U}\Lambda(v)(\tilde{R}_v - \tilde{U}^T V_v)$. Hence, differentiating (3) with respect to $V_v$ yields

$$\frac{2}{\sigma^2} \tilde{U}\Lambda(v)(\tilde{R}_v - \tilde{U}^T V_v) - \frac{2}{\sigma_V^2} V_v \quad (10)$$

Setting this equal to zero and rearranging yields (9). Similarly, differentiating (3) with respect to $\tilde{U}_g$ and setting equal to zero yields (8). □

It follows from Theorem 4.2 that to determine the matrices $\tilde{U}$, $V$ maximising $\log \tilde{p}(\tilde{U}, V | \mathcal{R}_\mathcal{O}; P)$ all that it is required for the recommender system to know is (i) the average ratings for each nym-item pair (namely, matrix $\tilde{R}$) and (ii) the number of users in each nym who rate item $v$ (namely, $\Lambda(v)$). There is no need for users to reveal their ratings in an individually-identifiable way.

Regarding the nym to user mapping $P$, observe that the log-posterior is separable in the columns of $P$,

$$\max_{P} \log p(\tilde{U}, V | \mathcal{R}_\mathcal{O}; P) = \min_{P} \sum_{(u,v) \in \mathcal{O}} (R_{uv} - P_u^T \tilde{U}^T V_v)^2$$
$$= \sum_{u \in \mathcal{U}} \min_{P_u} \sum_{v \in \mathcal{V}(u)} (R_{uv} - P_u^T \tilde{U}^T V_v)^2, \quad (11)$$

where $\mathcal{U} = \{u : (u,v) \in \mathcal{O}, v \in \{1, \cdots, m\}\}$ is the set of users providing ratings and, $\mathcal{V}(u) = \{v : (u,v) \in \mathcal{O}\}$ is the set of items rated by user $u$. Hence, we can find mapping $P$ for each user $u$ individually by solving $\min_{P_u} \sum_{v \in \mathcal{V}(u)} (R_{uv} - P_u^T \tilde{U}^T V_v)^2$. Provided the nym-item factorization matrices $\tilde{U}, V$ are made available to users by the recommender system, this optimisation can be carried out privately by each user on their own computer using their locally stored ratings $R_{uv}$. There is no need to release the individual user ratings $R_{uv}$ to the recommender system or to other users.

### 4.3. BLC Algorithm

The foregoing observations suggest an iterative algorithm that seeks to maximise the log posterior (1) by alternating between the following two steps:

(1) Using the current estimates for the matrix of average nym-item ratings $\tilde{R}$ and the number $\Lambda(v)$ of users in each nym who rate item $v$, estimate $\tilde{U}, V$ via Theorem 4.2.
(2) Given the current estimates for $\tilde{U}, V$ each user $u$ privately updates their column $P_u$ in $P$ by solving $\min_{P_u \in \mathcal{I}} \sum_{v \in \mathcal{V}(u)} (R_{uv} - P_u^T \tilde{U}^T V_v)^2$ where $\mathcal{I} = \{e_i, i = 1, p\}$, $e_i$ the vector for which element $i$ equals 1 and all other elements equal to 0. This optimisation can be trivially solved by simply calculating the objective for each element in (small) set $\mathcal{I}$ and selecting the lowest valued. The updated $\tilde{R}$ and $\Lambda(v)$ are then shared with the recommender system.

This two-stage process is given in more detail in Algorithm 1. At each iteration, each vector $\tilde{U}_g$ of nyms $g$ that are used by at least one user is updated using the current estimate of $V$. Then $V$ is updated using the current value of $\tilde{U}$. This is repeated until the improvement in the log-posterior falls below tolerance $\epsilon$. The users then update their choice of nym privately in a distributed manner and $\tilde{R}, \Lambda(v)$ are then updated accordingly.

### 4.4. Convergence

Since each update in Algorithm 1 is either a descent step or minimisation step, the algorithm is convergent and, in particular, we have:

THEOREM 4.3 (CONVERGENCE OF BLC ALGORITHM). *Suppose $\tilde{U}$ and $V$ remain bounded. Then the sequence generated by the alternating updates* (8) *and* (11) *convergences to a point of* $-\log \tilde{p}(\tilde{U}, V | \mathcal{R}_\mathcal{O}; P)$ *that is stationary for $\tilde{U}$ and $V$ and a local minimum for $P$.*

PROOF. *See Appendix.* □

### 4.5. Discussion

*4.5.1. Privacy-Preserving Calculation of $\tilde{R}, \Lambda(v)$.* Once users have updated the columns of $P$, the updated values of $\tilde{R}, \Lambda(v)$ need to be shared with the recommender system.

---

**Algorithm 1** BLC Nym Recommender
**loop**
    Initialise $E, \Delta \leftarrow \infty$, $\tilde{U} \leftarrow \mathcal{N}(0,\sigma)$, $V \leftarrow \mathcal{N}(0,\sigma)$
    **while** $E\Delta > \epsilon$ **do** {Block 1. System factorization}
        **for** $g = 1, \ldots, p$ **do**
            **if** $\sum_{v \in \mathcal{V}} \Lambda(v)_{gg} > 0$ **then**
$$\tilde{U}_g^T \leftarrow \sum_{v \in \mathcal{V}} \Lambda(v)_{gg} \tilde{R}_{gv} V_v^T \left( \frac{\sigma^2}{\sigma_{\tilde{U}}^2} I + \sum_{w \in \mathcal{V}} \Lambda(w)_{gg} V_w V_w^T \right)^{-1}$$
            **end if**
        **end for**
        **for all** $v \in \mathcal{V}$ **do**
$$V_v^T \leftarrow \tilde{R}_v^T \Lambda(v) \tilde{U}^T \left( \frac{\sigma^2}{\sigma_V^2} I + \tilde{U} \Lambda(v) \tilde{U}^T \right)^{-1}$$
        **end for**
        $\hat{E} \leftarrow \log \tilde{p}(\tilde{U}, V | \mathcal{R}_\mathcal{O}; P)$
        $\Delta \leftarrow |\hat{E} - E|$, $E \leftarrow \hat{E}$
    **end while**
    **for all** $u \in \mathcal{U}$ **do** {Block 2. User private nym choice}
        $P_u \leftarrow \min_{X_u \in \mathcal{I}} \sum_{v \in \mathcal{V}(u)} (R_{uv} - X_u^T \tilde{U}^T V_v)^2$
        **for all** $v \in \mathcal{V}(u)$ **do**
            $\Lambda(v) = \sum_{u \in \mathcal{U}(v)} P_u P_u^T$
        **end for**
        $\tilde{R}_v = \Lambda^{-1}(v) \sum_{u \in \mathcal{U}(v)} R_{uv} P_u \forall v \in \mathcal{V}$
    **end for**
**end loop**

---

This can be readily carried out in a privacy preserving manner and without the need for communication amongst the users.

To see this, consider first a simplified batch update where at each iteration users resubmit their ratings using their current choice of nym. Assuming users access the system via an appropriate anonymising connection, e.g. via Tor [Dingledine et al. 2004], then this update need not reveal the user identities. Given these ratings, the matrix $\tilde{R}$ of average item ratings by each nym can be immediately determined by the recommender system. The number of users sharing each nym can also be estimated, e.g. by counting the number of ratings for each item (assuming that each user submits at most one rating for each item, multiple ratings for an item provide an indication of the number of users sharing the nym). The extension of this to online submission of ratings is now straightforward. This setup is illustrated schematically in Figure 3.

*4.5.2. Scalability.* The BLC algorithm has favourable characteristics from the point of view of scalability. Firstly, the nym-item rating matrix $\tilde{R}$ which needs to be factorized is of dimension $p \times m$. In comparison, the user-item rating matrix $R$ is of dimension $n \times m$. Hence, the work scales with the number of nyms $p$ rather than the number of users $n$ and we expect $p \ll n$ (as we show in Section 5, in real scenarios $n > 70\,000$ and $p < 200$). Secondly, the nym-user mapping matrix $P$ is of dimension $p \times n$ but each column is updated by a separate user in a fully distributed manner, with no message passing or other co-ordination required between users. Hence, the update of $P$ is of the "embarrassingly parallel" type [Srirama et al. 2012] and so highly scalable.

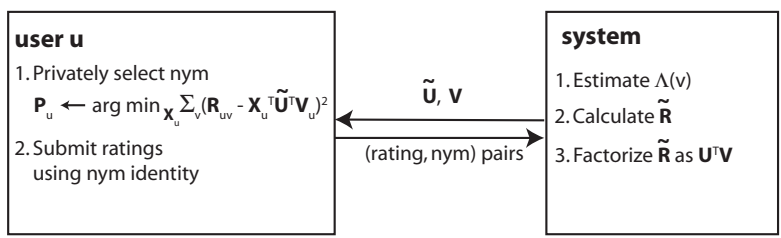

Fig. 3. Illustrating partitioning of operations in Algorithm 1 into a private user component and a public system component, with exchange only of nym-related information between both (no exchange of individually-identifying user data).

## 5. EXPERIMENTS USING SYNTHETIC DATA

We begin by using synthetic data sets where, by construction, we know ground truth regarding the number of nyms and so can evaluate the performance of the BLC algorithm against this. The setup considered in this section is intentionally tightly controlled so that we can vary one aspect at a time and study its impact. The performance with real ratings data will be considered later. We compare BLC with the classic matrix factorization algorithm [Koren et al. 2009].

### 5.1. Synthetic Data

To generate a data set with $p$ user groups we i) select $p$ points in the $d$ dimensional feature space as "group centres", and ii) randomly draw $n/p$ simulated users for each group, with position drawn from a Gaussian distribution with mean equal to the group position, and variable standard deviation. For instance, with $p = 5$ and a standard deviation equal to $0.01$, the users are tightly clustered around each of the group centres as illustrated in Figure 4a. We generate item matrix $V$ by drawing its elements from a Gaussian distribution, and thus obtain the full user-item matrix $R$ as $U^T V$. We then remove entries u.a.r. from $R$ to reach a desired degree of sparsity.

### 5.2. Performance vs Number of Nyms

Figure 4b shows the Root Mean Square Error (RMSE) of the predicted ratings obtained using the $P, \tilde{U}, V$ matrix factorization found by the BLC algorithm vs the number of nyms. The data shown is the average over 50 datasets drawn randomly as described above and error bars indicating one standard deviation are indicated. In this first example there are $m = 100$ items, the latent feature space is of dimension $d = 4$, there are $n = 10\,000$ users, $p = 5$ nyms and the users are clustered with standard deviation, $10^{-4}$ around the nym centres (i.e. tightly bunched) and $50\,\%$ of the user-item ratings are missing values. It can be seen that as the number of nyms used by the BLC algorithm is increased there is a sharp reduction in the RMSE once at least 5 nyms are used. Of course this is not surprising, but illustrates the ability of the BLC algorithm to correctly infer nym structure from the user-item rating data.

Also shown in Figure 4b is the RMSE of the predicted ratings when classic matrix factorization is used (where $P$ is fixed to be the $n \times n$ identity matrix). It can be seen that once the number of nyms used is 5 or greater, the RMSE of the BLC approach is consistently lower than that of the classic matrix factorization approach. That is, by taking advantage of the nym structure within the user ratings the BLC approach is able to make more accurate predictions since users sharing the same nym can leverage each others ratings.

Figure 5a plots the RMSE of the BLC predicted ratings vs the sparsity of the observed ratings matrix and the number $n$ of users. The data shown is the average over

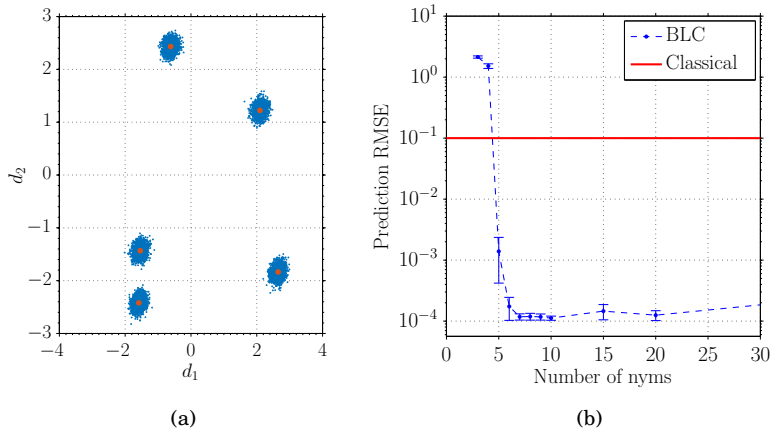

Fig. 4. (a) 2D projection of the positions in the $d = 4$ dimensional latent space of the users (in blue) and nyms (in red) (b) Prediction RMSE of BLC and classic matrix factorization vs. number of nyms.

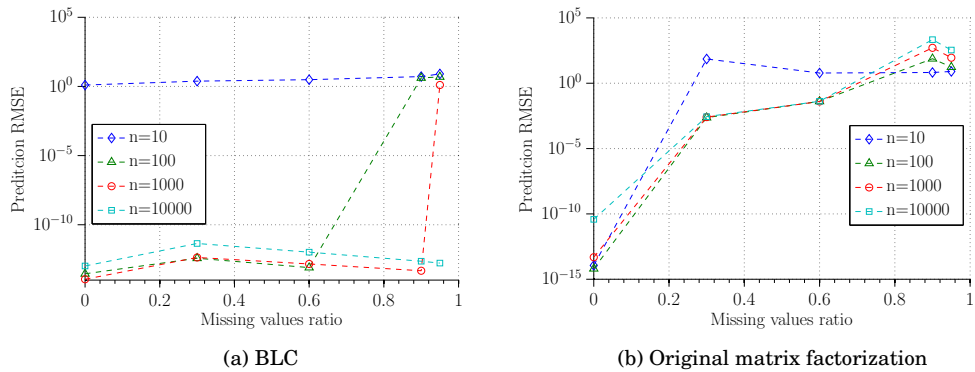

Fig. 5. Prediction RMSE vs. missing values ratio for BLC (a) with 5 groups and standard matrix factorization (b), for different number of users.

100 randomly drawn datasets and error bars are omitted because their size is negligible. It can be seen that as the number of users increases the prediction accuracy improves when the rating matrix is sparse. For $n = 1000$ or more users the prediction accuracy is insensitive to the fraction of missing values until this fraction exceeds 95 %. The poorer performance with smaller numbers of users is due to the reduced statistical multiplexing: the probability that an item has no ratings by any member of a nym becomes significant for sparse matrices with only a small number of users. Conversely, when there are larger numbers of users the nym structure allows users sharing a nym to leverage each others ratings to improve prediction accuracy when the ratings data is sparse.

Figure 5b plots the corresponding results when using the classic matrix factorization approach. It can be seen that for all numbers of users the prediction accuracy decreases as the observed ratings become more sparse. This approach cannot exploit the nym structure to improve predictions when the user data is sparse.

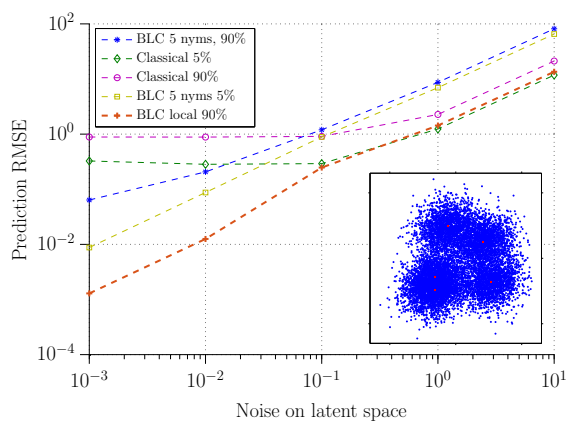

Fig. 6. RMSE of the predicted ratings vs. the standard deviation of users around the nym centres in the latent feature space, $n = 10\,000$ users, $p = 5$, $m = 100$, $d = 4$ and fraction of missing ratings $5\,\%$ and $90\,\%$.

### 5.3. Challenging the Nym Assumption

In the previous section the users were tightly bunched around the nym centres in the latent feature space. We now relax this and explore the impact of increasing the standard deviation and so the spread around the centres. Figure 6 shows the RMSE of the predicted ratings vs. the spread of the users around the nym centres. The inset plot in Figure 6 shows a 2D projection of the positions in the latent space of the users (in blue) and nyms (in red) for a standard deviation of $10$ about the nym centres. It can be seen that the prediction error with BLC increases roughly linearly with the standard deviation. The prediction accuracy is also largely insensitive to the sparsity of the ratings. For comparison, Figure 6 also shows the performance when the BLC algorithm is augmented using the extensions described in Sections 7.1 and 7.2. These extensions introduce adaptation of the number of nyms used, the number selected being shown in Figure 7. Note that as the standard deviation around the centres in the latent space increases we are degrading the original nym structure and so effectively increasing the number of groups and this is reflected in Figure 7. It can be seen that these extensions improve prediction performance, but do not change the qualitative behaviour, namely that the prediction error increases roughly linearly with the standard deviation around the centres in the latent space.

Also shown in Figure 6 is the corresponding data for the classic matrix factorization approach. For smaller values of standard deviation, where the data has a strong nym structure, it can be seen that BLC outperforms the classic approach in terms of prediction accuracy. However, as the standard deviation becomes large, so that the nym structure in the ratings is washed out, the prediction accuracy of the classic approach is better than that of the BLC approach unless the extensions in Sections 7.1 and 7.2 are used.

### 6. MEMORY AND TIME FOOTPRINT

To establish whether the proposed technique is practical and scalable, $100$ random scenarios were generated for a variety of combinations of $n, m$ values using 5 clusters with Gaussian noise as explained in Section 5. BLC algorithm performance is measured on a single core of an Intel Xeon Processor E3-1270 v3, $3.50\,\text{GHz}$. The overhead necessary to employ the per-routine measuring of memory and time consumption is significant, affecting the measured values by up to one order of magnitude. Nevertheless, this in-strumentation is important for understanding how memory and time consumption are affected by the different parameters of the algorithm.

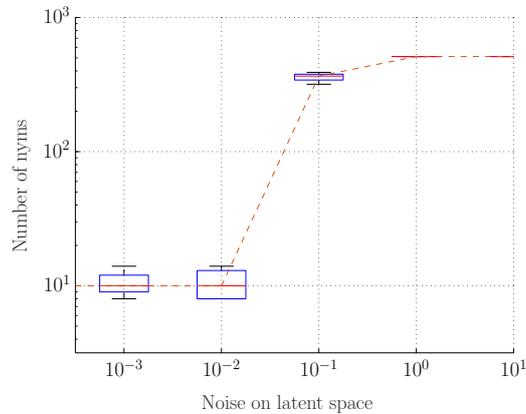

Fig. 7. Boxplot and median showing the number of nyms corresponding to the $BLC_{local}$ data in Figure 6.

In Figure 8 an analysis of the time and memory footprint vs $n$ and $m$ is shown. The error bars are not shown when the standard deviation is negligible. While memory usage increases linearly with both number of users and items, the convergence time increases linearly only with the number items, and remains roughly constant as the number of users is varied.

Importantly, we also analyse the scalability and performance when employing a warm start, i.e. the re-convergence performance with new but similar data drawn from the same population after initial convergence. In Figure 9(a) the convergence time to adjust to new data is analysed. The time to factorize a new group of users grows linearly with the number of users or items, but it is reasonably small ($6\,\text{s}$ for $1000$ users and $500$ items). Figure 9(b) shows the median time to run a matrix factorisation when up to $1\,\%$ of new users are added or when $1\,\%$ of existing ones change their rating: at worst the time to re-factorise is only around $60\,\text{ms}$. Regarding scalability, from Figure 10 it is clear that parallelisation has the potential to bring substantial benefits, for example if the factorization could be run in parallel client-side, since the convergence time per-user tends to a constant value.

## 7. EXTENSIONS TO BLC ALGORITHM

### 7.1. Selecting Number of Nyms

While BLC is designed to converge starting from any configuration of nyms in the latent space, in practice it may converge to a local optimum where some nyms stay unused or where two nyms could be coalesced into one to improve performance. We have found that a useful approach is therefore to initialise BLC with only 1 nym, then after convergence we double the number of nyms, adding Gaussian noise to the new nym feature vectors with standard deviation equal to half of the minimum inter-nym distance. At each doubling, nyms that are unused are pruned. In this way we essentially encourage users to progressively split into smaller groups where this leads to a performance improvement. We repeat this procedure until the factorization error falls below a threshold.

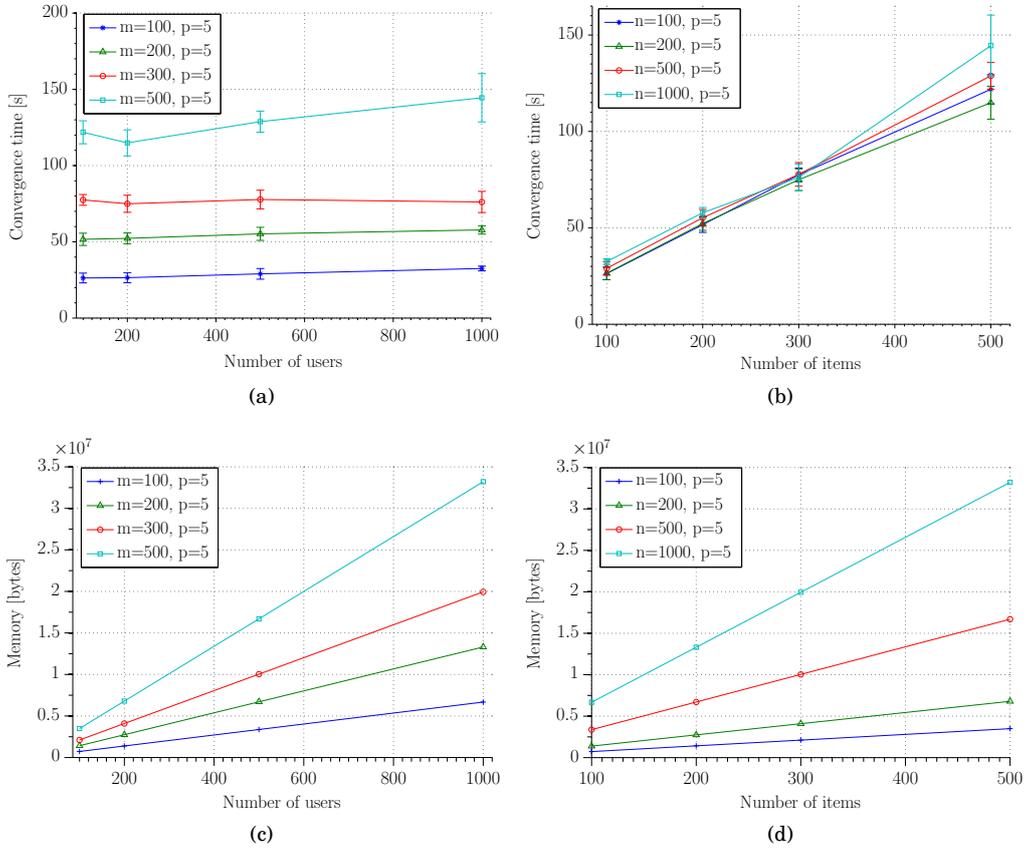

Fig. 8. Convergence time and memory usage vs number of users and number of items, for a synthetic scenario with 5 clusters.

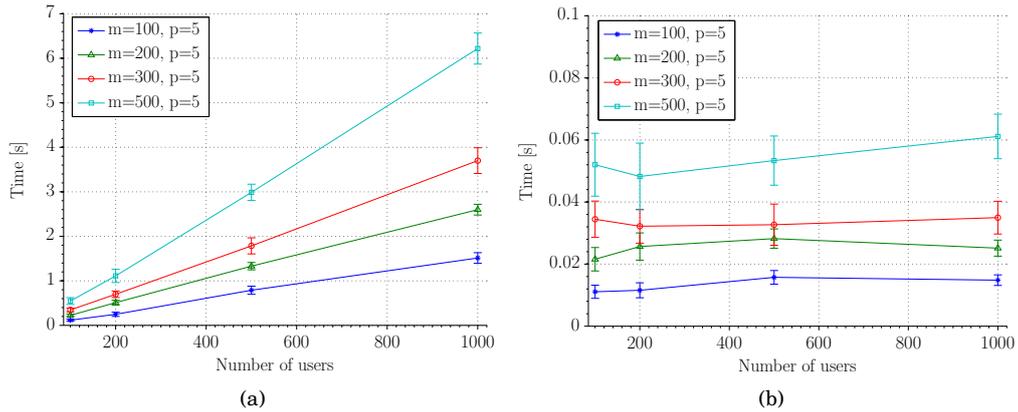

Fig. 9. Convergence time for (a) a full iteration after warm start, and (b) a single factorisation after up to 1 percent of ratings change.

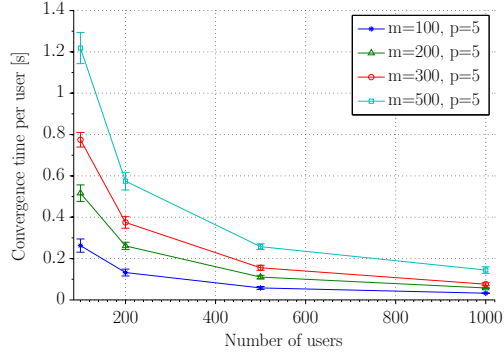

Fig. 10. Convergence time normalised by number of users vs number of users.

### 7.2. Local Recommendation

When users have access to the item profiles $V$, this allows increased flexibility in the way in which user ratings are predicted. For example, one approach is to use the predicted ratings $(\tilde{U}P_u)^T V$ for the nym selected by user $u$. Alternatively, user $u$ might exploit the information in their sole possession, i.e. the knowledge of their own private rating, to find the point $\hat{x}$ in the latent space that minimises $\sum_{v \in \mathcal{V}(u)} (R_{u,v} - \hat{x}^T V_v)^2$, given $V$ and the ratings $R_{u,\mathcal{V}(u)}$ already made by user $u$. This least squares optimisation can be solved locally by the user on their own computer, and so in a private manner. When the user has only rated a small number of items, this prediction is likely to be worse than the nym-based recommendation. However, we can combine the best from both worlds by solving a joint weighted least squared problem [Draper et al. 1966], in which user $u$ locates their position in the latent space that also minimise the distance from $\tilde{U}_{\cdot,g}$, where $g$ is the nym used by user $u$. More formally,

$$\hat{x} = \begin{bmatrix} R_{u,\mathcal{V}(u)} & \tilde{U}_g^T \end{bmatrix} \begin{bmatrix} I_{|\mathcal{V}(u)|} & 0 \\ 0 & wI_d \end{bmatrix} \begin{bmatrix} V_{\mathcal{V}(u)} & I_d \end{bmatrix}^T$$
$$\left( \frac{1}{\sigma_L^2} I_d + \begin{bmatrix} V_{\mathcal{V}(u)} & I_d \end{bmatrix} \begin{bmatrix} I_{|\mathcal{V}(u)|} & 0 \\ 0 & wI_d \end{bmatrix} \begin{bmatrix} V_{\mathcal{V}(u)} & I_d \end{bmatrix}^T \right)^{-1},$$

where $I_k$ is a diagonal $k \times k$ matrix and $w, \sigma_L^2$ are design parameters. Then the recommendation is computed by $\hat{x}V$. We will refer to this enhanced algorithm by $\text{BLC}_{\text{local}}$. We test this technique in section 8.

### 7.3. Factorization Frequency

As illustrated in Figure 3, Algorithm 1 is partitioned into a private component run by each user $u \in \mathcal{U}$ and a public system component. Given (nym,rating) pairs supplied by the users the system component calculates nym rating matrix $\tilde{R}$ and factorizes it into $\tilde{U}, V$ using an estimate of $\Lambda(\cdot)$. The user component run by user $u$ selects the nym to be used by user $u$ using the current value of $\tilde{U}, V$. Rather than factorizing $\tilde{R}$ every time a new rating is supplied or a user changes nym, the system workload can be reduced by carrying out the factorization less frequently. Figure 11 plots the prediction RMSE vs. the factorization frequency and the number of users $n$ over 500 samples, using the same parameters as in the other simulations (but without Gaussian noise). The update frequency is specified as the fraction of users that have executed their private part of Algorithm 1 between each factorization, and a subset of users performs updates between each matrix factorization (user ordering is chosen from a permutation drawn uniformly at random for every execution of Block 2 of Algorithm 1). It can be seen that, perhaps unsurprisingly, when we re-factorize after every individual user update we obtain the best performance, and performance decreases as the factorization frequency decreases. That is, there is a trade-off between computational effort and prediction accuracy. However, as the number of users $n$ increases it can be seen that the performance cost becomes less, presumably because the absolute number of user updates between factorizations is increasing, and for $n > 1000$ users a factorization period of $10\,\%$ seems reasonable.

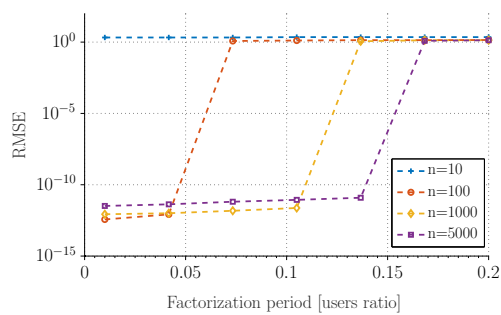

Fig. 11. Prediction RMSE vs. factorization frequency.

Table I. Dataset parameters.

| Dataset | Users | Items | Ratings | Density | Domain |
|---|---|---|---|---|---|
| Jester | 73 421 | 100 | 4.1 M | 0.5584 | $(-10, 10)$ |
| Movielens | 71 567 | 10 681 | 10 M | 0.0131 | $(1, 5)$ |
| Dating | 135 359 | 168 791 | 17.3 M | 0.0007 | $(1, 10)$ |
| Books | 278 858 | 271 379 | 1.1 M | 0.00001 | $(1, 10)$ |
| Netflix | 17 770 | 480 189 | 1.4 M | 0.0001 | $(1, 5)$ |

### 7.4. Cold Start

In all of the experiments, we simulate the problem of *cold start*, by starting with no ratings and gradually filling the set $\mathcal{O}$ of observed ratings one user at a time. We cycle through a random permutation of the users until $\mathcal{O}$ contains the full training set. Every time a new user is added, all of their ratings are added to $\mathcal{O}$. After the first permutation of users is completed, every time a user is selected it will potentially change $\tilde{R}, P$ and $\Lambda(\cdot)$, moving all its ratings to a different nym.

## 8. PERFORMANCE WITH REAL DATA

### 8.1. Jester, Movielens, Dating & Books Datasets

We follow the workflow and use the results of [Kannan et al. 2014]: the algorithms compared are ALSWR (alternating least squares with regularisation) [Zhou et al. 2008], SGD [Funk 2006], SVD++ [Koren 2008], Bias-SVD [Koren 2008] and BMF [Kannan et al. 2014] on the datasets summarised in Table I.

For each dataset we used the latent space dimension $d$ that yielded the best RMSE in [Kannan et al. 2014], and used the same proportional split between training, validation and test data ($85\,\%$, $5\,\%$ and $10\,\%$ respectively) and the same validation technique (subsampling validation). We include the Book dataset for completeness of comparison, even if it is a sort of "degenerate" dataset, so sparse that using the global book average has been proven to be more effective that the state-of-the-art collaborative filtering techniques. We stress that in the following only the BMF algorithm has optimal performance because the other algorithms have many other parameters (other than $d$) that could be optimised, but that have been kept at their default values by the authors of the comparison. We did not tune the BLC regularisation parameters, instead keeping the variances of the priors large (1000) and relying on the averaging between the users sharing the same nym to control overfitting.

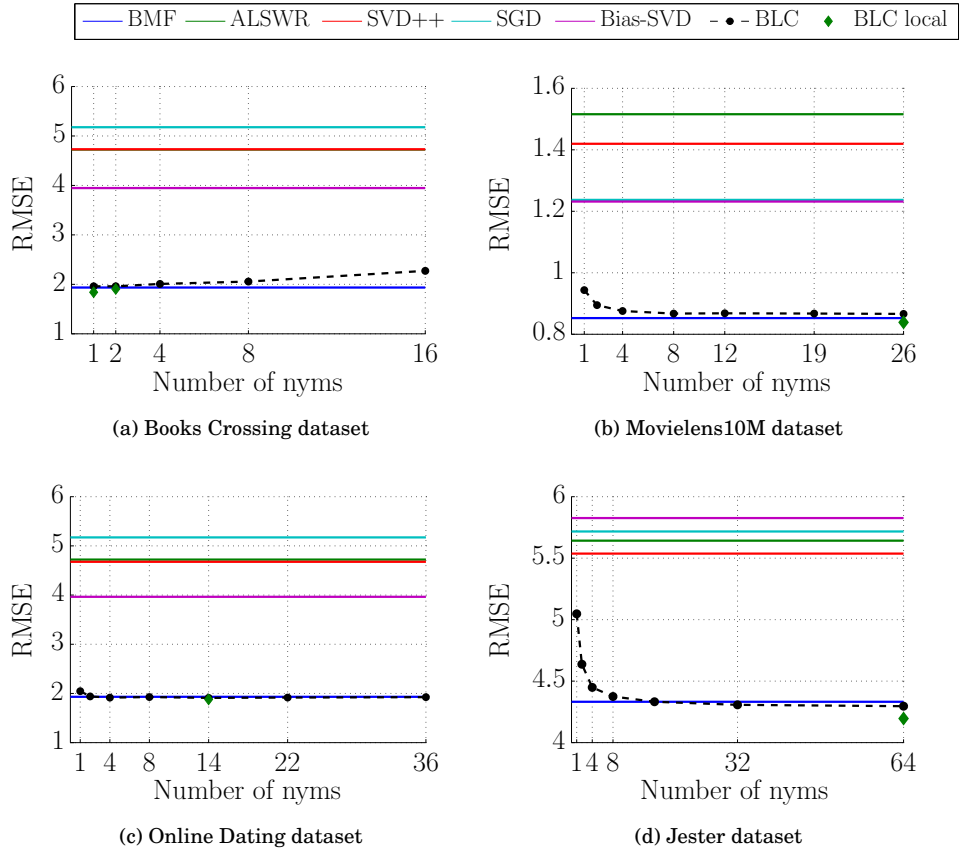

Fig. 12. Prediction RMSE vs number of nyms used compared with state-of-the-art algorithms.

Figure 12 shows the RMSE of BLC vs. the number of nyms used and Table II summarises the RMSE values. It can be seen that BLC and $BLC_{local}$ are competitive with the state of the art, with only minimal tuning (namely of the number of nyms). That is, privacy-enhanced recommendation need not come at the cost of reduced prediction accuracy.

Note also that only a small number of nyms is needed: fewer than 20 nyms gives good performance for all datasets. That is, real data does indeed appear to possess a group structure of the kind we have assumed. Further, the small number of nyms means that we can expect that each nym is shared by a large number of users thereby providing a strong hiding in the crowd form of privacy – this is discussed in more detail below.

Table II. Summary of the RMSE performance for the various algorithms and datasets considered, using validation sets from [Kannan et al. 2014].

| **Dataset** | BMF | ALSWR | SVD++ | SGD | Bias SVD | **BLC** | **BLC local** | (nyms) |
|---|---|---|---|---|---|---|---|---|
| Jester | 4.33 | 5.64 | 5.54 | 5.72 | 5.82 | 4.30 | **4.20** | 64 |
| Movielens | 0.85 | 1.51 | 1.42 | 1.24 | 1.23 | 0.87 | **0.83** | 26 |
| Dating | 1.93 | 4.72 | 4.68 | 5.17 | 3.96 | 1.91 | **1.88** | 14 |
| Books | 1.94 | 4.71 | 4.73 | 5.18 | 3.95 | 1.96 | **1.87** | 1 |

### 8.2. Another Comparison for Movielens 10M

We now compare BLC with another family of algorithms, based on nuclear norm regularisation, for the Movielens 10M dataset. The algorithms are as follows:

*JSH.* An extension of the Frank-Wolfe algorithm for optimising a function over the bounded positive semidefinite cone [Jaggi et al. 2010].
*Soft-Impute.* A soft singular value thresholding algorithm [Mazumder et al. 2010].
*SSGD-Matrix-Completion.* A more advanced algorithm based on stochastic subgradient descent [Avron et al. 2012].
*GECO.* A greedy method with optimality guarantees for low rank matrix factorization [Shalev-Shwartz et al. 2011].

For all of the algorithms, we report the RMSE from [Avron et al. 2012] in Table III. Also shown in this table is the median RMSE of BLC over 5 samples, using the same parameters of before and converging at $8$ nyms. It can be seen that, once again, the prediction accuracy of BLC is competitive with, if not superior to, the state of the art.

Table III. Performance comparison of algorithms for the Movielens 10M dataset, using validation sets from [Avron et al. 2012].

| SSGD | JSH | Soft-Impute | GECO | **BLC** | **BLC$_{local}$** |
|---|---|---|---|---|---|
| 0.8555 | 0.8640 | 0.8605 | 0.8771 | 0.8720 | **0.8452** |

### 8.3. Netflix Dataset

Our evaluation would not be complete without showing results for the classic Netflix challenge datasets, for which all the algorithms listed above have been initially designed. Unfortunately, the Netflix validation set is no longer available. Following [Kannan et al. 2014], we therefore tested BLC against the validation set that was supplied as part of the training package (using the same $1.4\,\mathrm{M}$ ratings training set described in Table I). As before, we used the latent space dimension $d = 20$ that yielded the best RMSE for the other algorithms in [Kannan et al. 2014] and we did not tune BLC parameters. It can be seen from Table IV that BLC achieves the second best performance, using 128 nyms.

Table IV. Performance comparison of algorithms for the Netflix dataset, BLC paramters were untuned.

| BMF | ALSWR | SVD++ | SGD | Bias-SVD | Time-svd | **BLC** | **BLC$_{local}$** |
|---|---|---|---|---|---|---|---|
| 0.9533 | 1.5663 | 1.5453 | 1.2997 | 1.3882 | 1.1829 | **0.9875** | **0.9785** |

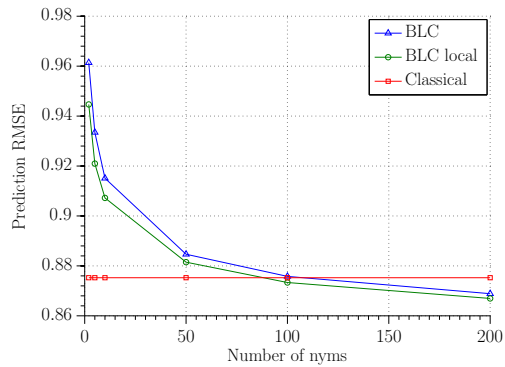

Fig. 13. RMSE vs number of nyms used for Netflix dataset.

In Figure 13, we compare BLC with the classic matrix factorization algorithm for a dense ($26\,\%$) subset of the Netflix dataset, obtained by selecting the top $1000$ items and the top $10\,000$ users that rated them. These results are useful to further confirm that use of a small number $p = 100$ of nyms is sufficient to match the classic matrix factorization approach on this dataset, and for a larger number of nyms the performance surpasses that of the classic approach.

## 9. PRIVACY PERFORMANCE

As discussed in Section 3, the attacks of interest can be decomposed into two types: (i) attacks which seek to learn which nym a user belongs to, and (ii) attacks which seek to learn which items a user has rated, and what that rating is, given knowledge of the nym to which the user belongs.

With regard to (i), lacking additional side information the attacker can try to guess the correct nym using the information in $\Lambda$, i.e. exploiting the fact that some nyms might contain more users than others: the prevalent nym is the best bet for the attacker, and the more the user-nym distribution is far from the uniform distribution, the more chance the attacker has to correctly guess the nym used by an arbitrary user. In the worst case for the attacker (uniform distribution) the user can get an indistinguishability probability of $1/p$, where $p$ is the number of nyms i.e. we have a form of $k$-anonymity [Sweeney 2002] with $k = p$. In general, the probability $P_g$ of guessing the right nym is equal to the number of users in the largest nym divided by the total number of users. Table V plots $P_g$ for each of the datasets considered in the previous section, with Figure 14 showing more detail for the Movielens dataset. For the online book dataset BLC converged with 1 nym, leading to a probability of guessing $P_g$ of $100\,\%$. But it is interesting to note that if we let the algorithm run until it is using 8 nyms, $P_g$ decreases to $15.24\,\%$ with a mere increase of $1.2\,\%$ in the prediction RMSE. All the other datasets have a $P_g$ of less than $23\,\%$. The use of nyms therefore provides a reasonable level of protection against this first type of attack.

We now consider the second type of attack of interest, where an attacker seeks to learn which items a user has rated given knowledge of the nym to which the user belongs. We define the following privacy measure:

$$p(n,j) = \frac{\Lambda(j)_{nn}}{\sum_{v \in \mathcal{V}} \Lambda(v)_{nn}}.$$

Table V. Indistinguishability performance for BLC algorithm: number of nyms used and probability $P_g$ of guessing which nym contains a target user.

| **Dataset** | Jester | Movielens [Avron et al. 2012] | Dating | Book | Netflix |
|---|---|---|---|---|---|
| **no. nyms** | 64 | 8 | 14 | 1 (8) | 128 |
| $P_g[\%]$ | 3.5 | 22.17 | 19.4 | 100 (23.21) | 2.71 |

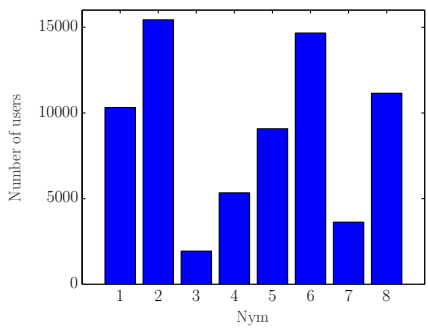

Fig. 14. Number of users per nym in Movielens dataset [Avron et al. 2012].

where $n$ is the nym index, $j$ the item index, $\Lambda(j)_{nn}$ the number of users in nym $n$ who have rated item $j$ and $\sum_{v \in \mathcal{V}} \Lambda(v)_{nn}$ is the total number of users sharing nym $n$. It can be seen that $p(n, j)$, which we refer to as the *association probability*, is an estimate of the probability that a user rated movie $j$, given that the user chose nym $n$. Hence, smaller values of $p(n, j)$ correspond to increased privacy in the sense that it is harder for an attacker to learn that a specific user in nym $n$ rated item $j$, whereas when $p(n, j) = 1$ then every user in nym $n$ has rated item $j$ and so an attacker can be certain that a target user in nym $n$ has rated that item. Intuitively, we expect that as the number of users increases then $p(n, j)$ will tend to decrease and so privacy increase. Figure 15 shows the distribution over all the nyms of the association probability of the worst item. As expected, it can be seen that there is a downward trend with the number of users for both the Movielens and online dating datasets. When the number of users is more than around 10% of that in the dataset then the association probability is always less than 0.3 i.e. each user can deny that they rated any given item with probability at least 0.7 even when an attacker knows which nym the user belongs to. This seems like a substantial level of deniability, sufficient for most practical purposes. Figure 16 shows the impact of increasing the number of nyms used. It can be seen that using more nyms improves prediction accuracy but reduces the level of privacy, as might be expected. However, the reduction in privacy is relatively small, with the association probability remaining consistently below 0.3.

Compared to the other algorithms considered in Section 8, BLC leads to an association probability decrease of at least 70% in all cases.

## 10. CONCLUSIONS

We propose a privacy-enhanced matrix factorization recommender that exploits the fact that users can often be grouped together by interest. This allows a form of "hiding in the crowd" privacy. We introduce a novel matrix factorization approach suited to making recommendations in a shared group (or nym) setting and the BLC algorithm for carrying out this matrix factorization in a privacy-enhanced manner. We demonstrate that the increased privacy does not come at the cost of reduced recommendation accuracy since, when nyms are chosen appropriately, users sharing a nym can leverage their shared ratings/preferences to make high quality recommendations.

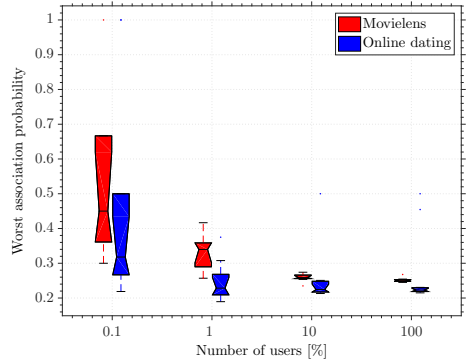

Fig. 15. Boxplot of the association probability of the worst item vs. the number of users in the system for the Movielens and Online dating datasets when using BLC.

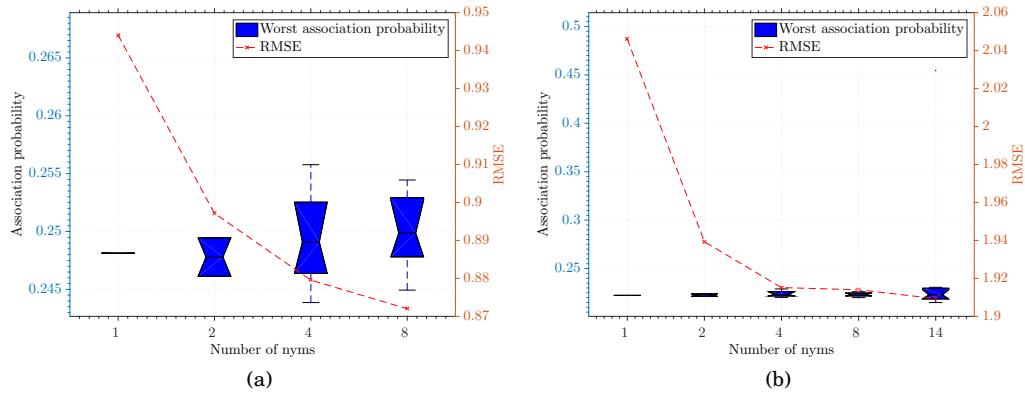

Fig. 16. Association probability (left y axis) of worst item and RMSE (right y axis) of BLC for Movielens (a) and Online dating (b) datasets. The distribution of the worst item association probability over the nyms is shown as a boxplot (left y axis) and the RMSE of the prediction is shown vs the number of nyms used by BLC.

**REFERENCES**

Charu C Aggarwal. 2005. On k-anonymity and the curse of dimensionality. In *Proceedings of the 31st international conference on Very large data bases*. VLDB Endowment, 901–909.

Rakesh Agrawal and Ramakrishnan Srikant. 2000. Privacy-preserving data mining. In *ACM Sigmod Record*, Vol. 29. ACM, 439–450.

Esma Aïmeur, Gilles Brassard, José M Fernandez, and Flavien Serge Mani Onana. 2008. ALAMBIC: a privacy-preserving recommender system for electronic commerce. *International Journal of Information Security* 7, 5 (2008), 307–334.

Haim Avron, Satyen Kale, Shiva Kasiviswanathan, and Vikas Sindhwani. 2012. Efficient and practical stochastic subgradient descent for nuclear norm regularization. *arXiv preprint arXiv:1206.6384* (2012).

Patrick Billingsley. 2008. *Probability and measure*. John Wiley & Sons.

Ludovico Boratto and Salvatore Carta. 2011. State-of-the-Art in Group Recommendation and New Approaches for Automatic Identification of Groups. In *Information Retrieval and Mining in Distributed Environments*. Studies in Computational Intelligence, Vol. 324. Springer, 1–20.

Joseph Calandrino, Ann Kilzer, Arvind Narayanan, Edward W Felten, Vitaly Shmatikov, and others. 2011. "You Might Also Like:" Privacy Risks of Collaborative Filtering. In *Security and Privacy (SP), 2011 IEEE Symposium on*. IEEE, 231–246.

John Canny. 2002. Collaborative filtering with privacy. In *Security and Privacy, 2002. Proceedings. 2002 IEEE Symposium on*. IEEE, 45–57.

David Chaum, Amos Fiat, and Moni Naor. 1990. Untraceable electronic cash. In *Proceedings on Advances in cryptology*. Springer-Verlag New York, Inc., 319–327.

Anupam Datta, Divya Sharma, and Arunesh Sinha. 2012. Provable de-anonymization of large datasets with sparse dimensions. In *Principles of Security and Trust*. Springer, 229–248.

Chris Ding, Tao Li, Wei Peng, and Haesun Park. 2006. Orthogonal nonnegative matrix t-factorizations for clustering. In *Proceedings of the 12th ACM SIGKDD international conference on Knowledge discovery and data mining*. ACM, 126–135.

Xuan Ding, Lan Zhang, Zhiguo Wan, and Ming Gu. 2010. A brief survey on de-anonymization attacks in online social networks. In *Computational Aspects of Social Networks (CASoN), 2010 International Conference on*. IEEE, 611–615.

Roger Dingledine, Nick Mathewson, and Paul Syverson. 2004. *Tor: The second-generation onion router*. Technical Report. DTIC Document.

Norman Richard Draper, Harry Smith, and Elizabeth Pownell. 1966. *Applied regression analysis*. Vol. 3. Wiley New York.

S Funk. 2006. Stochastic gradient descent. (2006). http://sifter.org/simon/journal/20061211.html

Jagadeesh Gorla, Neal Lathia, Stephen Robertson, and Jun Wang. 2013. Probabilistic Group Recommendation via Information Matching. In *22nd Int. Conf. on World Wide Web (WWW '13)*. 495–504.

Saikat Guha, Bin Cheng, and Paul Francis. 2011. Privad: Practical Privacy in Online Advertising. In *8th USENIX NSDI Symposium*.

Thomas Hofmann. 2004. Latent semantic models for collaborative filtering. *ACM Transactions on Information Systems (TOIS)* 22, 1 (2004), 89–115.

Zhengli Huang, Wenliang Du, and Biao Chen. 2005. Deriving private information from randomized data. In *Proceedings of the 2005 ACM SIGMOD international conference on Management of data*. ACM, 37–48.

Martin Jaggi, Marek Sulovsk, and others. 2010. A simple algorithm for nuclear norm regularized problems. In *Proceedings of the 27th International Conference on Machine Learning (ICML-10)*. 471–478.

Ramakrishnan Kannan, Mariya Ishteva, and Haesun Park. 2014. Bounded matrix factorization for recommender system. *Knowledge and information systems* 39, 3 (2014), 491–511.

Hillol Kargupta, Souptik Datta, Qi Wang, and Krishnamoorthy Sivakumar. 2003. On the privacy preserving properties of random data perturbation techniques. In *Data Mining, 2003. ICDM 2003. Third IEEE International Conference on*. IEEE, 99–106.

Michal Kompan and Maria Bielikova. 2013. Group Recommendations: Survey and Perspectives. *Computing and Informatics* 33, 2 (2013), 446–476.

Yehuda Koren. 2008. Factorization meets the neighborhood: a multifaceted collaborative filtering model. In *Proceedings of the 14th ACM SIGKDD international conference on Knowledge discovery and data mining*. ACM, 426–434.

Yehuda Koren, Robert Bell, and Chris Volinsky. 2009. Matrix factorization techniques for recommender systems. *Computer* 8 (2009), 30–37.

Shyong K Lam and John Riedl. 2004. Shilling recommender systems for fun and profit. In *Proceedings of the 13th international conference on World Wide Web*. ACM, 393–402.

Dongsheng Li, Qin Lv, Huanhuan Xia, Li Shang, Tun Lu, and Ning Gu. 2011. Pistis: a privacy-preserving content recommender system for online social communities. In *IEEE Int. Conf. on Web Intelligence and Intelligent Agent Technology*, Vol. 1. 79–86.

Rahul Mazumder, Trevor Hastie, and Robert Tibshirani. 2010. Spectral regularization algorithms for learning large incomplete matrices. *The Journal of Machine Learning Research* 11 (2010), 2287–2322.

Animesh Nandi, Armen Aghasaryan, and Makram Bouzid. 2011. P3: A privacy preserving personalization middleware for recommendation-based services. In *Hot Topics in Privacy Enhancing Technologies Symposium*.

Arvind Narayanan and Vitaly Shmatikov. 2006. How to break anonymity of the netflix prize dataset. *arXiv preprint cs/0610105* (2006).

Jennifer Nguyen and Mu Zhu. 2013. Content-boosted matrix factorization techniques for recommender systems. *Statistical Analysis and Data Mining* 6, 4 (2013), 286–301.

Valeria Nikolaenko, Stratis Ioannidis, Udi Weinsberg, Marc Joye, Nina Taft, and Dan Boneh. 2013. Privacy-preserving matrix factorization. In *ACM SIGSAC Computer & Communications Security (CCS'13)*.

Mark O'Connor, Dan Cosley, Joseph A Konstan, and John Riedl. 2001. PolyLens: a recommender system for groups of users. In *ECSCW 2001*. 199–218.

Huseyin Polat and Wenliang Du. 2005. SVD-based collaborative filtering with privacy. In *Proceedings of the 2005 ACM symposium on Applied computing*. ACM, 791–795.

Shai Shalev-Shwartz, Alon Gonen, and Ohad Shamir. 2011. Large-scale convex minimization with a low-rank constraint. *arXiv preprint arXiv:1106.1622* (2011).

Shang Shang, Pan Hui, Sanjeev R Kulkarni, and Paul W Cuff. 2011. Wisdom of the crowd: Incorporating social influence in recommendation models. In *17th IEEE Int. Conf. on Parallel and Distributed Systems (ICPADS)*.

R. Shokra, P. Pedarsani, G. Theodorakopoulos, and J.P. Hubaux. 2009. Preserving Privacy in Collaborative Filtering through Distributed Aggregation of Offline Profiles. In *Proc RecSys*.

K Shyong, Dan Frankowski, John Riedl, and others. 2006. Do you trust your recommendations? An exploration of security and privacy issues in recommender systems. In *Emerging Trends in Information and Communication Security*. Springer, 14–29.

Satish Narayana Srirama, Pelle Jakovits, and Eero Vainikko. 2012. Adapting scientific computing problems to clouds using MapReduce. *Future Generation Computer Systems* 28, 1 (2012), 184–192.

Latanya Sweeney. 2002. k-anonymity: A model for protecting privacy. *International Journal of Uncertainty, Fuzziness and Knowledge-Based Systems* 10, 05 (2002), 557–570.

Lyle H Ungar and Dean P Foster. 1998. Clustering methods for collaborative filtering. In *AAAI workshop on recommendation systems*, Vol. 1. 114–129.

Quan Wang, Zheng Cao, Jun Xu, and Hang Li. 2012. Group Matrix Factorization for Scalable Topic Modeling. In *35th Int. ACM Conf. on Research and Development in Information Retrieval (SIGIR '12)*. 375–384.

Yu Xin and Tommi Jaakkola. 2014. Controlling privacy in recommender systems. In *Advances in Neural Information Processing Systems*. 2618–2626.

Gui-Rong Xue, Chenxi Lin, Qiang Yang, WenSi Xi, Hua-Jun Zeng, Yong Yu, and Zheng Chen. 2005. Scalable collaborative filtering using cluster-based smoothing. In *Proceedings of the 28th annual international ACM SIGIR conference on Research and development in information retrieval*. ACM, 114–121.

Guangyou Zhou, Yubo Chen, Daojian Zeng, and Jun Zhao. 2014. Group Non-negative Matrix Factorization with Natural Categories for Question Retrieval in Community Question Answer Archives. In *COLING 2014, 25th Int. Conf. on Computational Linguistics: Technical Papers*.

Yunhong Zhou, Dennis Wilkinson, Robert Schreiber, and Rong Pan. 2008. Large-scale parallel collaborative filtering for the netflix prize. In *Algorithmic Aspects in Information and Management*. Springer, 337–348.

Kai Zhu, Rui Wu, Lei Ying, and R Srikant. 2014. Collaborative filtering with information-rich and information-sparse entities. *Machine Learning* 97, 1-2 (2014), 177–203.

PROOF OF THEOREM 4.3. We begin by observing that $tr\tilde{U}^T\tilde{U}$ is the Frobenius norm of $\tilde{U}$, and thus is convex in $\tilde{U}_u$ for all $u$ and similarly $tr V^T V$ is convex in $V_v$ for all $v$. We also have that $\sum_{v \in \mathcal{V}}(\tilde{R}_v - \tilde{U}^T V_v)^T \Lambda(v)(\tilde{R}_v - \tilde{U}^T V_v)$ is convex in $V_v$ (it is the usual least squares objective in $V_v$). Since the Hessian of this function w.r.t. $U_u$ is positive definite for all $u$, we also have that it is convex in $U_u$. It follows that $-\log \tilde{p}(\tilde{U}, V | \mathcal{R}_\mathcal{O}; P)$ is individually convex in $\tilde{U}$ and $V$.

Now consider the sequence $\tilde{U}^{(k)}$, $V^{(k+1)}$, $k = 1, 2, \ldots$ generated by alternating updates

$$(\tilde{U}_g^{(k)})^T = \sum_{v \in \mathcal{V}} \Lambda(v)_{gg} \tilde{R}_{gv}(V_v^{(k)})^T \left( \frac{\sigma^2}{\sigma_{\tilde{U}}^2} I + \sum_{w \in \mathcal{V}} \Lambda(w)_{gg} (V_w^{(k)} V_w^{(k)})^T \right)^{-1} \quad (12)$$

$$(V_v^{(k+1)})^T = \tilde{R}_v^T \Lambda(v)(\tilde{U}^{(k)})^T \left( \frac{\sigma^2}{\sigma_V^2} I + \tilde{U}\Lambda(v)(\tilde{U}^{(k)})^T \right)^{-1} \quad (13)$$

By Theorem 4.2, $\tilde{U}^{(k)}$ is a stationary point of $-\log \tilde{p}(\cdot, V^{(k)} | \mathcal{R}_\mathcal{O}; P)$ and by convexity it is therefore a global minimum of $-\log \tilde{p}(\tilde{U}, V | \mathcal{R}_\mathcal{O}; P)$ when $V$ and $P$ are held fixed.

A:25Similarly, $\boldsymbol{V}^{(k)}$ is a stationary point and so global minimum of $-\log \tilde{p}(\tilde{\boldsymbol{U}}, \cdot | \mathcal{R}_\mathcal{O}; \boldsymbol{P})$ when $\tilde{\boldsymbol{U}}$ and $\boldsymbol{P}$ are held fixed. It follows that the sequence of log posteriors satisfies $-\log \tilde{p}(\tilde{\boldsymbol{U}}^{(k)}, \boldsymbol{V}^{(k)} | \mathcal{R}_\mathcal{O}; \boldsymbol{P}) \geq -\log \tilde{p}(\tilde{\boldsymbol{U}}^{(k)}, \boldsymbol{V}^{(k+1)} | \mathcal{R}_\mathcal{O}; \boldsymbol{P}) \geq -\log \tilde{p}(\tilde{\boldsymbol{U}}^{(k+1)}, \boldsymbol{V}^{(k+1)} | \mathcal{R}_\mathcal{O}; \boldsymbol{P})$, $k = 1, 2, \ldots$, that is, it is descending.

When $\boldsymbol{U}$, $\boldsymbol{V}$ and $\boldsymbol{P}_u$ are held fixed for all $u \neq \hat{u}$, the update (11) for variable $\hat{u}$ is a global minimum of $-\log p(\tilde{\boldsymbol{U}}, \boldsymbol{V} | \mathcal{R}_\mathcal{O}; \boldsymbol{P})$ and, by Lemma 4.1, a global minimum of $-\log \tilde{p}(\tilde{\boldsymbol{U}}, \boldsymbol{V} | \mathcal{R}_\mathcal{O}; \boldsymbol{P})$. It follows that sequence $-\log \tilde{p}(\tilde{\boldsymbol{U}}^{(k)}, \boldsymbol{V}^{(k)} | \mathcal{R}_\mathcal{O}; \boldsymbol{P}^{(k)})$, $k = 1, 2, \ldots$ is also decreasing.

Matrix $\boldsymbol{P}^k$ is, by construction, $(0,1)$ valued and so uniformly bounded for all $k$. Assuming that $\tilde{\boldsymbol{U}}^{(k)}$, $\boldsymbol{V}^{(k+1)}$ also remain uniformly bounded for all $k$ then by the monotone convergence of bounded sequences (e.g. [Billingsley 2008, Theorem 16.2]), sequence $\log \tilde{p}(\tilde{\boldsymbol{U}}^{(k)}, \boldsymbol{V}^{(k)} | \mathcal{R}_\mathcal{O}; \boldsymbol{P})$ converges to a finite limit as $k \to \infty$. Let $\log \tilde{p}_\infty$ denote this limit and let $C_\infty$ denote the corresponding set of limit points.

It remains to show that every limit point is a stationary point for $\tilde{\boldsymbol{U}}$ and $\boldsymbol{V}$ and a local minimum for $\boldsymbol{P}$. For $\tilde{\boldsymbol{U}}$ and $\boldsymbol{V}$ this follows from the alternating update and Theorem 4.2. □